# Envelope imbalanced ensemble model with deep sample learning and local-global structure consistency

Fan Li, Xiaoheng Zhang, Yongming Li*, Pin Wang

School of Microelectronics and Communication Engineering, Chongqing University, Chongqing, China 400044

% Thanks are due to Dr Yinhua Shen for valuable advice to this paper.

## Abstract

The class imbalance problem is important and challenging. Ensemble approaches are widely used to tackle this problem because of their effectiveness. However, existing ensemble methods are always applied into original samples, while not considering the structure information among original samples. The limitation will prevent the imbalanced learning from being better. Besides, research shows that the structure information among samples includes local and global structure information. Based on the analysis above, an imbalanced ensemble algorithm with the deep sample pre-envelope network (DSEN) and local-global structure consistency mechanism (LGSCM) is proposed here to solve the problem. This algorithm can guarantee high-quality deep envelope samples for considering the local manifold and global structures information, which is helpful for imbalance learning. First, the deep sample envelope pre-network (DSEN) is designed to mine structure information among samples. Then, the local manifold structure metric (LMSM) and global structure distribution metric (GSDM) are designed to construct LGSCM to enhance distribution consistency of interlayer samples. Next, the DSEN and LGSCM are put together to form the final deep sample envelope network –DSEN-LG. After that, base classifiers are applied on the layers of deep samples respectively. Finally, the predictive results from base classifiers are fused through bagging ensemble learning mechanism. To demonstrate the effectiveness of the proposed method, forty-four public datasets and more than ten representative relevant algorithms are chosen for verification. The experimental results show that the algorithm is significantly better than other imbalanced ensemble algorithms.

## 1.Introduction

Class imbalance problem is widely present in many fields of data analysis and mining. So far, it has been one of the most hotly researched and challenging problems. It refers to datasets with different sample sizes in different categories. When class imbalance is encountered, the classifier usually favors the majority class and therefore cannot correctly classify the minority class, resulting in performance loss. Classifier may treat some samples in minority classes as outliers, resulting in biased and inaccurate classification models [1]. With the increasing size of datasets in a lot of real-world applications, such as target identification [2], business analytics [3], anomaly detection [4], and medical diagnosis [5], the class imbalance problem has become more serious in their consequences.

The existing methods for solving the class imbalance problem are mainly categorized into data level methods, algorithm level methods, ensemble methods and feature selection methods. These methods can tackle the class imbalance problem from different aspects. The purpose of the data level methods is to rebalance the prior distribution of classes during preprocessing, including oversampling for minority classes, undersampling for majority classes, and a combination of both [6-8]. Algorithm level methods use different misclassification costs, allowing the minority class to be more focused such as cost-sensitive learning [9]. Ensemble methods incorporate ensemble learning and data level or algorithm level methods [10-12]. For example, data-level ensemble methods aim to combine different resampling strategies with bagging and boosting (e.g., (RUSBoost) [13], (SMOTEBoost) [14], Underbagging[15] and Overbagging[16]). The feature selection methods aim to choose a feature subset that are better suited to the class imbalance problem [17].

Ensemble learning is used extensively to address the class imbalance problem because of its effectiveness in adaptability, robustness, generalization, and their superior performance over single classifier [18-21]. Its main idea is to combine several complementary classifiers to enhance the single classifier. The ensemble learning methods are not initially designed to tackle the class imbalance problem, but their relative robustness and superior performance enabled them to gain popularity. To achieve optimal performance for class imbalance problem, it is necessary to incorporate ensemble learning methods with data-level or algorithm-level methods [22].

Most ensemble methods modify the sampling step (i.e., data-level) in ensemble learning such as bagging and boosting. In addition, some algorithms modify the misclassification cost (i.e., algorithm level) when training and combining ensemble members, but cost-sensitive ensembles are less popular in practical use, because it is difficult to determine the appropriate costs for a specific task [23-24]. On the contrary, combining data-level methods with ensemble learning is relatively simple due to the independence between the ensemble training phases and data sampling [22]. It has shown that data-level ensembles provide an alternative to solve class imbalance problems. For example, SMOTE, which is representative oversampling methods, can be combined with the bagging and boosting, including SMOTEBoost [14], SMOTEBagging [18]. Besides, to improve the quality of generated data, Generative Adversarial Networks (GANs) can also be used for generating minority class samples [25]. Undersampling methods are also considered such as UnderBagging[15], BalancedBagging [26], EasyEnsemble and BalanceCascade [27].Hybrid data-level methods can also be embedded into bagging and boosting frameworks. In addition to simply combining the ensemble with data-level approaches, recent studies have provided effective methods about how to design better data-level ensemble solutions. Sun et al.[28] first split an imbalanced dataset into several balanced clusters, and trained each base classifier on the basis of the balanced data. The adopted strategy uses the balanced dataset to train each base classifier and can enhance the diversity within the ensemble. Lim et al.[29] design an ensemble framework with cluster-based synthetic oversampling. Wang et al.[30] design an ensemble framework with a new SMOTE method. The new SMOTE method considers the boundary information of samples. Kang et al. [31] design a classifier ensemble framework with an undersampling strategy. The undersampling strategy mainly eliminates noise in minority class through a noise filter, as noise degrades classifier performance. Chih-Fong Tsai et al. [32] combine a certain sampling approach with ensemble classifiers. The sampling approach first clusters the majority class and then selects samples from each cluster. Liu et al.[33] design a novel ensemble framework with an undersampling strategy. This strategy can self-paced harmonize data hardness via undersampling to generate a robust ensemble. Kaixiang Yang et al.[34] propose a hybrid optimal ensemble classifier framework with a density-based undersampling method. This undersampling method aims to select informative sample through probabilistic-based data transformation, obtaining balanced subsets. Zhi Chen et al. [35] develop an ensemble framework with a hybrid data-level strategy. This method combines ensemble learning with the diversity-enhancing oversampling and union of a margin-based undersampling.

During data collection, samples are often taken from the same subject or from the same community, so these samples have hierarchical structural information. Taking speech diagnosis as example, multiple corpora are collected for the same subject, and some subjects are from same community, so the corpora samples with hierarchical structure information are used for classification [36-37]. Existing data-level ensemble methods only consider modeling on the original samples, so they have difficulty in reflecting the hierarchical relationship between samples. Therefore, these ensemble methods may suffer from the limitation, since the structure information of samples are not considered during modeling.

To address the above issues, we need to study the new data-level ensemble method which can explore the structure information of samples and construct new representative samples for imbalance learning. Deep learning is currently high-quality feature learning method and hot spot. The big possible reason for its success is in that deep neural networks can extract features with high-quality through multilayer transformations, thereby exploring structure information of original features and obtaining new representative features. Inspired by the principle of the deep learning, multilayer clustering of samples can be designed for exploring structure information of original samples and obtaining high quality of new samples. The representative new samples better characterize the structure information and category properties of original samples. Since the category properties of the minority and majority samples are 1:1, these new samples will help to achieve a balancing of the different categories of the samples. It is worth noting that multilayer clustering algorithms have been proposed for classifying in recent years [38-39]. These studies have shown that multilayer clustering can improve clustering performance, as it can mine the structure information of samples and obtain new representative samples. However, existing multilayer clustering methods aim to cluster, not to generate new samples to be classified. Moreover, the number of clustering centers is approximated to the number of clusters, instead of the samples. Besides, these clustering methods do not take into account the distribution of interlayer samples, etc. Therefore, by combining the ideas of deep learning and multilayer clustering, it is necessary to design deep clustering method for generation of new high-quality samples.

As to the clustering algorithm, existing clustering algorithms include the density peak clustering, Fuzzy C-Means (FCM), K-means, Self-Organizing Map (SOM), Spectral clustering, etc. FCM, as a soft clustering algorithm, is often used, and was proposed by Bezek [40]. Soft clustering algorithm is characterized by allowing each sample to be a member of many clusters to some extent. The degree of membership indicates the possibility of belonging to a cluster, with a value between 0 and 1, and the membership value determines which cluster the sample belongs to. Samples can be more informative and diverse as they do not directly belong to a single clustering center. So, in this paper, FCM is selected as the clustering algorithm to implement sample transformation. As mentioned above, to explore more sample information, it is necessary to design deep clustering by combining multilayer clustering and deep transformation. Notably, multilayer clustering is different from the existing ones. The former is for generation of new samples, while the latter is for direct classification.

Although multilayer clustering can be used for sample transformation to obtain more information, there are often inconsistencies in the distribution between interlayer samples, so this affects the quality of sample generation (transformation). To further solve this problem, we will consider to explore the structure information of interlayer samples, thereby guaranteeing the consistency of interlayer samples. The structure information of the sample contains local manifold and global structures information [41]. Research showed that, considering both local manifold and global structures information achieved good results in domain adaptation [42]. Therefore, we will explore local manifold and global structures of the sample. As the global distributions discrepancy can be measured by Maximum Mean Discrepancy (MMD) in domain adaptation [43] and the local structure distributions discrepancy can be measured by manifold learning [42], the MMD and manifold learning are combined for exploring local and global structure information in this proposed method.

Based on the above ideas, this paper proposes an imbalanced ensemble algorithm based on the deep sample envelope network (DSEN-LG). First, the deep sample envelope pre-network (DSEN) is designed to mine structure information among samples, including sample neighborhood concatenation (SNC) and deep envelope sample generation based on multilayer FCM (MlFCM). Then, the LMSM and GSDM are designed to construct LGSCM to enhance distribution consistency of interlayer samples. Next, the DSEN and LGSCM are put together to form the final deep sample envelope network –DSEN-LG. After that, base classifiers are applied on the layers of deep samples respectively. Finally, the predictive results from base classifiers are fused through bagging ensemble learning mechanism to get the final result.

The main contributions of this paper are as follows.

1) Existing imbalanced ensemble methods are always applied into original samples, while not considering the structure information among samples. This paper proposes a deep sample envelope pre-network (DSEN) to mine the structure information among samples through multilayer sample transformation.

2) For the multi-layer clustering, there exists distribution inconsistency of interlayer samples, but existing clustering methods ignore the problem. This paper proposes a local-global structure consistency mechanism (LGSCM) to fully address this problem by consider the local manifold and global structures information of the interlayer samples.

3) Different from existing imbalanced ensemble methods, this paper proposes an imbalanced ensemble algorithm based on DSEN and LGSCM, to explore the local and global structure information of original samples, thereby constructing high quality of new samples and obtaining satisfied accuracy.

The rest of this paper is structured as follows: The proposed algorithm is presented in detail in Section 2; Section 3 gives the experiment results and analysis; and the discussion and conclusion are provided in Section 4.

## 2.Methods

The proposed imbalanced ensemble algorithm mainly combines data-level approach and bagging ensemble learning approach. The algorithm realizes the deep envelope sample generation based on the network DSEN-LG and solves the imbalance problem based on the generated deep envelope samples. The proposed network DSEN-LG mainly contains DSEN and local-global structure consistency mechanism (LGSCM). DSEN aims to mine the samples' structure information and LGSCM aims to enhance consistency of interlayer deep envelope samples. Table 1 lists the key symbols, and Table 2 shows the relevant terms in this paper.

Table 1. Key symbols

| Symbols | Definition |
| --- | --- |
| $NN_K$ | The selected k-nearest neighbors |
| $V_e^L$ | The $L$th-layer deep envelope samples generated by DSEN |
| $V_e'^L$ | The $L$th-layer deep envelope samples generated by DSEN-LG |
| $C_L$ | The $L$th-layer number of clusters |
| $\varphi$ | An implicit but generic transformation |
| $\Psi$ | Kernel Gram matrix |
| $\oplus$ | Concatenation operator |
| $\lfloor \ \rfloor$ | Rounding down |

Table 2. Relevant terms

| Terms | Definition |
| --- | --- |
| DSEN | Deep sample envelope pre-network |
| SNC | Sample neighborhood concatenation |
| LGSCM | Local-global structure consistency mechanism |
| DSEN-LG | The network combining the DSEN and LGSCM |
| DSEN-LGIE | The proposed imbalanced ensemble learning algorithm |
| D&F | Division and fusion |

## 2.1. Deep Sample Envelope Pre-Network (DSEN)

To mine the original samples' structure information, the deep sample envelope pre-network (DSEN) is proposed. Firstly, the complementary information between samples is enhanced through sample neighborhood concatenation (SNC), and then the deep envelope samples are obtained through MlFCM.

### 2.1.1. Sample Neighborhood Concatenation

Consider a dataset $X = \{x_1, x_2, ... x_n\} \in \mathbb{R}^{n \times s}$ with $n$ samples and $s$ features. For the sample $x, x \in X$, the k-nearest neighbors of the sample can be found using the Euclidean distance metric through following equation

$$d(x, x_i) = \sqrt{(x - x_i)^T (x - x_i)}, 1 \leq i \leq n \tag{1}$$

Let $NN_K(x) = \{nn_x^i | nn_x^i \in X\}_{i=1}^{K}$ denote the selected k-nearest neighbors of $x$, where $K \leq n$, and through Eq.(1), an ascending list of the distance between $x$ and $NN_K(x)$ can be displayed as follow:

$$d(x, nn_x^1) \leq d(x, nn_x^2) \leq ... \leq d(x, nn_x^K) \tag{2}$$

The ranking of $d(x, nn_x^i)$ in the distance sequence $DS = \{d(x, nn_x^i) | nn_x^i \in X\}_{i=1}^{K}$ can simply represent the level of similarity between $x$ and $nn_x^i$. So we can define the k-nearest neighbors search more formally in the following manner [44].

$$NN_K(x, X, K) = A \tag{3}$$

Where $A$ is a set satisfying the conditions as follow,

$$A \subseteq X, \forall x_p \in A, x_q \in X - A, \ d(x, x_p) \leq d(x, x_q) \tag{4}$$

The k-nearest neighbors sample set $A$ is obtained and concatenated with the original sample $x$ to form an envelope sample $x_e$

$$x_e = x \oplus A \tag{5}$$

$\oplus$ is defined the concatenation operator. So, from Eq(5), the envelope dataset $X_e = \{x_{1e}, x_{2e}, ..., x_{ne}\} \in R^{n \times (K+1)s}$ can be obtained based on the original dataset $X \in R^{n \times s}$ transformation.

### 2.1.2. Deep envelope sample generation

Suppose $X_e = \{x_{1e}, x_{2e}, ..., x_{ne}\} \in R^{n \times (K+1)s}$ denotes original dataset, $V_e = \{v_1, v_2, .., v_c\} \in R^{c \times (K+1)s}$ denotes the corresponding prototypes of $C$ clusters. FCM can be used to cluster $X_e$ by minimizing.

$$\min J(U, V_e) = \sum_{i=1}^{c} \sum_{p=1}^{n} u_{ip}^m d_{ip}^2, \ \text{s.t.} \sum_{i=1}^{c} u_{ip} = 1 \tag{6}$$

where $d_{ip} = \|x_{pe} - v_i\|$ represents Euclidean distance, $u_{ip}$ is the membership degree of $x_{pe}$ to the $i$th cluster, partition matrix $U = (u_{ip})_{c \times n}$. $m$ denotes fuzzification coefficient ($m > 1$) and it is usually set to 2. Through minimizing Eq.(6), the partition matrix and prototype iteration formulas can be obtained as follow.

$$u_{ip} = \frac{1}{\sum_{j=1}^{c} \left[\frac{d_{ip}}{d_{jp}}\right]^{\frac{2}{m-1}}}, \ v_i = \frac{\sum_{p=1}^{n} (u_{ip})^m x_{pe}}{\sum_{p=1}^{n} (u_{ip})^m} \tag{7}$$

Through FCM clustering, the original dataset $X_e$ can be transformed into $V_e^1 = \{v_1, v_2, .., v_{c_1}\} \in R^{c_1 \times (K+1)s}$ by Eq.(7), and $V_e^1$ as a new dataset can also be transformed into $V_e^2 = \{v_1, v_2, .., v_{c_2}\} \in R^{c_2 \times (K+1)s}$ by Eq.(7). If $L$-layer sample transformation based on FCM is performed, the new sample points $V_e^L = \{v_1, v_2, .., v_{c_L}\} \in R^{c_L \times (K+1)s}$ can be obtained. So the original dataset can be transformed through multilayer clustering. We refer to this method as Multilayer Fuzzy C-Mean Clustering (MlFCM) and MlFCM is achieved on the basis of single layer FCM(SlFCM). The deep envelope samples $V_e^L$ can be obtained by MlFCM.

The DSEN combines the SNC and MlFCM. The overall scheme is shown in Figure 1. Figure 1(a) denotes the SNC module, Figure 1(b) denotes the SlFCM module and Figure 1(c) denotes MlFCM. The pseudocode description for DSEN is shown in Algorithm 1.

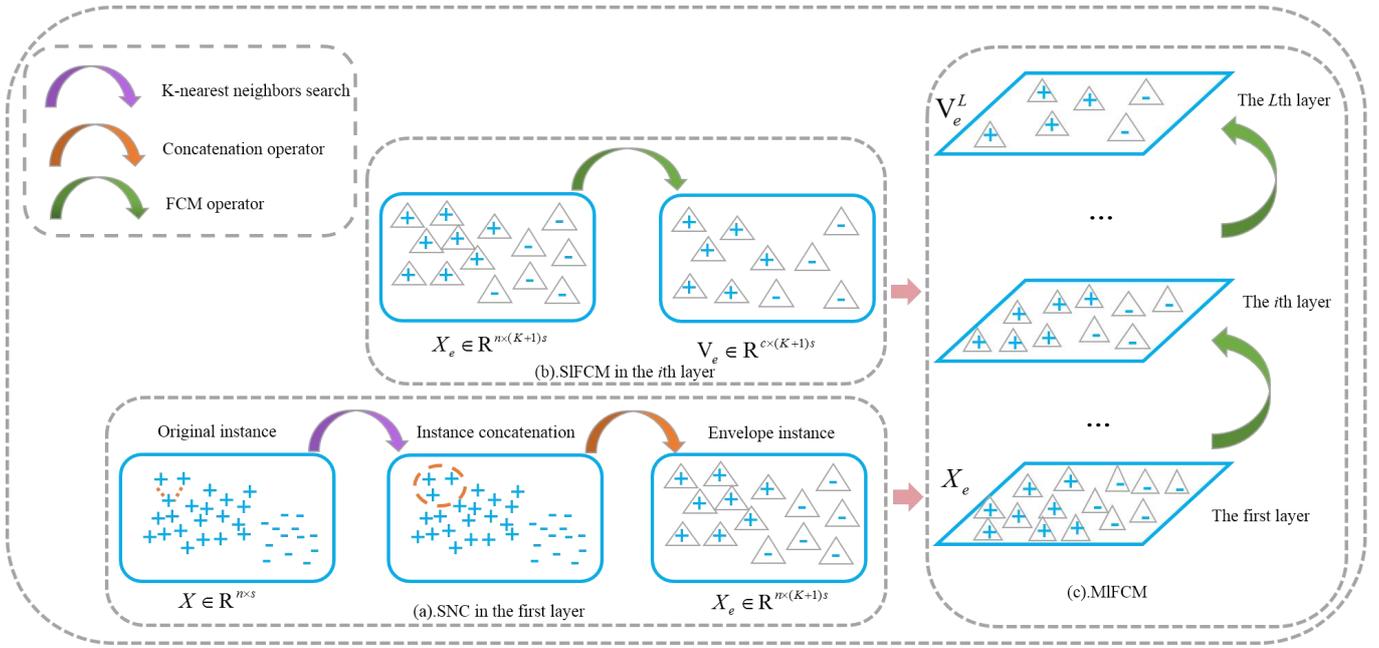

Fig.1. Deep Sample Envelope Pre-Network (DSEN): (a)SNC, (b)SlFCM, (c)MlFCM

---

**Algorithm 1:** Deep Sample Envelope Pre-Network-DSEN

**Input:** Original dataset $X = \{x_1, x_2, \ldots x_n\}$, the number of clusters per layer $C_1, \ldots, C_L$, the number of layers of clustering $L$, the number of nearest neighbor samples $K$; fuzzification coefficient $m$, iteration number $w$ and threshold $\varepsilon$.

**Output:** Generated deep envelope samples $V_e^L$.

**Procedure:**

1. Find the nearest neighbor samples $A$ for the original sample via (1);

2. Concatenated the original sample with $A$ to form the envelope dataset $X_e$ via (5);

3. For $l = 1 : L$

4.    Initialize the degree of membership matrix $U$ randomly;

5.    $w \leftarrow 1$;

6.    **Repeat**

7.    Update new samples $V_e^{l(w)} = \{v_1, v_2, \ldots, v_{c_l}\}$ via (7);

8.    $w \leftarrow w + 1$;

9.    **Until** $\left| J(U, V_e)^{(w+1)} - J(U, V_e)^{(w)} \right| < \varepsilon$;

10.   **Return** $V_e^l$ and $V_e^l$ is used as the input of the next layer;

11. End

12. Return the deep envelope samples $V_e^L$.

---

## 2.2. Local-global structure consistency mechanism (LGSCM)

The interlayer deep envelope samples can be obtained based on DSEN. In order to enhance consistency of interlayer deep envelope samples, the local-global structure consistency mechanism (LGSCM) is proposed here. This LGSCM aims to explore more structure information of the samples including local manifold and global structures. Specifically, $X_e, V_e$ denote the interlayer deep envelope sample sets in the DSEN. Through a projection matrix $P$, the datasets $X_e, V_e$ are mapped to a potential common subspace and a matrix $\mathcal{G}$ is introduced to generate a transition dataset $X_M$ in the common subspace. $P$ maps the data space from $\mathrm{R}^{(K+1)s}$ to a latent subspace $\mathrm{R}^d$ ($(K+1)s \geq d$). By minimizing the local and global distribution between $X_M$ and $X_e$, the interlayer data distribution in the DSEN is made consistent.

### 2.2.1. Local manifold structure metric (LMSM)

LMSM is proposed to enhance the consistency of local structure distribution between $X_e$ and $V_e$ indirectly, by constraining the generative transition dataset $X_M$. So local structure preservation of this paper can be defined as follow

$$\mathcal{L}_{\mathrm{LMSM}(F_{X_M}, F_{X_e})} = \sum_{h,e}^{n} S_{hk} \|\varphi(x_h) - \varphi(x_e)\|_2^2 = Tr\left(\varphi(X_M) D\left(\varphi(X_M)\right)^\mathrm{T}\right)$$
$$+ Tr\left(\varphi(X_e) D \left(\varphi(X_e)\right)^\mathrm{T}\right) \quad (8)$$
$$- 2 Tr\left(\varphi(X_M) S \left(\varphi(X_e)\right)^\mathrm{T}\right)$$

Where $F_{X_M}$ and $F_{X_e}$ denote the distribution of $X_M$ and $X_e$, respectively. $\varphi(x_h) \in \varphi(X_M) = \varphi(V_e)\mathcal{G}$, $\varphi(x_e) \in \varphi(X_e)$. $Tr(\bullet)$ is the matrix trace and $\varphi$ indicates an implicit but generic transformation. The diagonal matrix $D$ is a matrix with entries $D_{hh} = \sum_{e=1}^{n} S_{he}$ and $S$ is the affinity matrix, calculated as

$$S_{he} = \begin{cases} 1, & \text{if } x_h \in NN_K(x_e) \| x_e \in NN_K(x_h) \\ 0, & \text{otherwise} \end{cases} \quad (9)$$

Let $P^\mathrm{T} = \Theta^\mathrm{T} \varphi(X_r)^\mathrm{T}$, $X_r = [V_e, X_e]$, the projected $X_e, V_e$ can be denoted as $\Theta^\mathrm{T} \varphi(X_r)^\mathrm{T} \varphi(X_e)$ and $\Theta^\mathrm{T} \varphi(X_r)^\mathrm{T} \varphi(V_e)$. So after projection, the Eq.(8) can be further expressed as

$$\min_{\Theta, \mathcal{G}} \frac{1}{n^2} Tr\left(\Theta^\mathrm{T} \Psi_v \mathcal{G} D \left(\Theta^\mathrm{T} \Psi_v \mathcal{G}\right)^\mathrm{T}\right)$$
$$+ \frac{1}{n^2} Tr\left(\Theta^\mathrm{T} \Psi_e D \left(\Theta^\mathrm{T} \Psi_e\right)^\mathrm{T}\right) \quad (10)$$
$$- \frac{2}{n^2} Tr\left(\Theta^\mathrm{T} \Psi_v \mathcal{G} S \left(\Theta^\mathrm{T} \Psi_e\right)^\mathrm{T}\right)$$

Where $\Psi_V = \varphi(X_r)^\mathrm{T} \varphi(V_e)$, $\Psi_e = \varphi(X_r)^\mathrm{T} \varphi(X_e)$ are kernel matrices.

## 2.2.2. Global structure distribution metric (GSDM)

GSDM is proposed to reduce the discrepancy of global distribution between the $X_e$ and $V_e$ indirectly, by constraining the generative transition dataset $X_M$. So the GSDM can be expressed as follow

$$\mathcal{L}_{\text{GSDM}(F_{X_M}, F_{X_e})} = \frac{1}{n} \sum_{h=e=1}^{n} \|\varphi(x_h) - \varphi(x_e)\|_2^2 \tag{11}$$

Similarly, after projection, the Eq. (11) can be further expressed as

$$\min_{\Theta, \mathcal{G}} \frac{1}{n} \|\Theta^T (\Psi_v \mathcal{G} - \Psi_e) \mathbf{1}\|_2^2 \tag{12}$$

where **1** represents a column vector with element one.

## 2.2.3. Joint optimization of LMSM and GSDM

The proposed LGSCM aims to align the distribution of interlayer deep envelope samples and preserve local manifold structure. Therefore, through the projection matrix $P$, in the common subspace, the local and global distribution discrepancy between $X_e$ and $V_e$ can be minimized by combining LMSM(10), GSDM (12) and low rank constraint(LRC) regularization [45] as follow

$$\begin{aligned}
\min_{\Theta, \mathcal{G}} & \frac{1}{n^2} Tr\left(\Theta^T \Psi_v \mathcal{G} D (\Theta^T \Psi_v \mathcal{G})^T\right) \\
& + \frac{1}{n^2} Tr\left(\Theta^T \Psi_e D (\Theta^T \Psi_e)^T\right) \\
& - \frac{2}{n^2} Tr\left(\Theta^T \Psi_v \mathcal{G} S (\Theta^T \Psi_e)^T\right) + \frac{\lambda}{n} \|\Theta^T (\Psi_v \mathcal{G} - \Psi_e) \mathbf{1}\|_2^2 + \lambda_1 \|\mathcal{G}\|_* \\
s.t \quad & \Theta^T \Psi \Theta = I
\end{aligned} \tag{13}$$

Where $\Psi = \varphi(X_r)^T \varphi(X_r)$ is kernel matrix and the nuclear norm $\|\mathcal{G}\|_*$ is the low rank constraint. $\lambda, \lambda_1$ are tradeoff parameters.

The problem (13) can be written with the augmented Lagrange function by introducing an auxiliary variable $\mathcal{H}$ as follow,

$$\begin{aligned}
\min_{\Theta, \mathcal{G}, \mathcal{H}} & \frac{1}{n^2} (Tr\left(\Theta^T \Psi_v \mathcal{G} D (\Phi^T \Psi_v \mathcal{G})^T\right) + Tr\left(\Theta^T \Psi_e D (\Theta^T \Psi_e)^T\right) - 2Tr\left(\Theta^T \Psi_v \mathcal{G} S (\Theta^T \Psi_e)^T\right)) \\
& + \frac{\lambda}{n^2} \Theta^T (\Psi_v \mathcal{G} \mathbf{1} (\Psi_v \mathcal{G})^T - \Psi_v \mathcal{G} \mathbf{1} (\Psi_e)^T - \Psi_e \mathbf{1} (\Psi_v \mathcal{G})^T + \Psi_e \mathbf{1} (\Psi_e)^T) \Theta + \lambda_1 \|\mathcal{H}\|_* \\
& + Tr\left(\zeta_1^T (\mathcal{G} - \mathcal{H})\right) + \frac{\delta}{2} \left(\|\mathcal{G} - \mathcal{H}\|_F^2\right)
\end{aligned} \tag{14}$$

Where $\zeta_1$ is the LaGrange multiplier, and $\delta$ is a penalty parameter. The optimization process of the three variables $\Theta, \mathcal{H}, \mathcal{G}$ can be used by following [42]. When optimal $\Theta$ is obtained, the new deep envelope sample set is obtained by $V_e' = \Theta^T \Psi_V$. The overall scheme of LGSCM is portrayed in Figure 2 and the pseudocode description for LGSCM is shown in Algorithm 2. $V_e'$ is obtained and then input the next layer of DSEN. After *L*-layer DSEN, the final deep envelope sample can be obtained $V_e'^L$. So the DSEN and LGSCM are combined to form a new network called DSEN-LG and the deep envelope sample $V_e'^L$ generated via DSEN-LG for tackling the class imbalance problem.

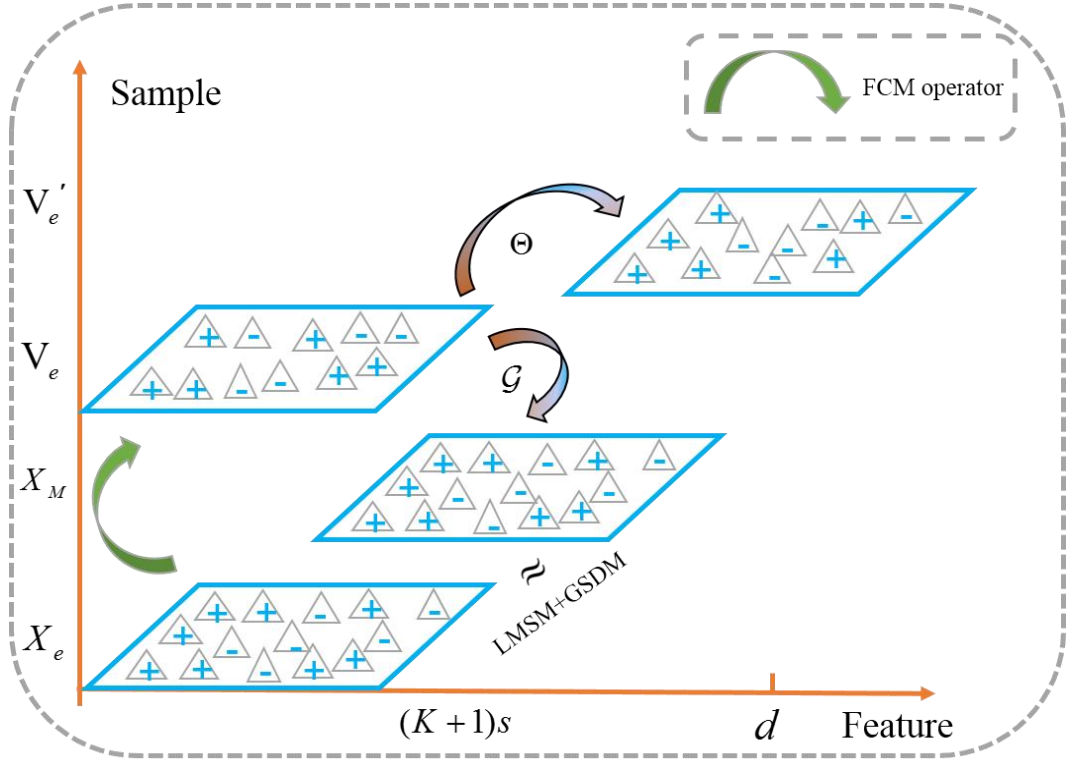

Fig.2. Local-global structure consistency mechanism (LGSCM)

---

**Algorithm 2**: Local-global structure consistency mechanism (LGSCM)

**Input:** Interlayer deep envelope sample sets $X_e, V_e$

**Output:** Generated deep envelope sample set $V_e'$

**Procedure:**

1.Initialize: $\mathcal{H} = \mathcal{G} = 0$;

2.Compute $\Psi = \varphi(X_r)^T \varphi(X_r)$, $\Psi_V = \varphi(X_r)^T \varphi(V_e)$, $\Psi_e = \varphi(X_r)^T \varphi(X_e)$, $X_r = [V_e, X_e]$;

3.Optimize $\Theta, \mathcal{H}, \mathcal{G}$ and compute $V_e' = \Theta^T \Psi_V$ when $\Theta$ is optimal;

4.Return $V_e'$.

---

## 2.3. Whole method

An imbalanced ensemble algorithm based on the deep sample envelope network - DSEN-LG (called DSEN-LGIE) was proposed. In this proposed algorithm, firstly, the majority class samples in the imbalanced training set will be divided into $Q$ subsets, and each subset contains a number of majority class samples equal to minority class samples in the training set. We perform subset partitioning based on feature weighting [46], rather than random sampling or clustering. Assuming $n_1, n_2$ denote the number of majority samples $X_{maj}(maj=1,2,...,n_1) \in R^{n_1 \times s}$ and minority samples $X_{min} \in R^{n_2 \times s}$ in the imbalanced training set respectively, the feature-weighted sum of each sample in the majority class can be calculated as follows.

$$y = \sum_{f=1}^{s} w_{majf} x_{majf}, \quad w_{majf} = \frac{x_{majf}}{\sum_{f=1}^{s} x_{majf}} \tag{15}$$

Where $x_{majf}$ denotes the value of the $f$th feature for $x_{maj}$, $w_{majf}$ denotes the weight of the $f$th feature for $x_{maj}$. Each sample is given an index value $y$ by Eq.(15). The majority class samples are sorted in ascending order according to the corresponding ascending order of $y$, and then divide the sorted majority class sample into $Q$ subsets. Each subset contains $n_2$ majority class samples. Specifically, the ordered samples of the 1st to the $n_2$th could be obtained as the 1st subset; the $n_2+1$th to the $2n_2$th samples are obtained as the second subset, etc. A balanced training set can be obtained by fusing the majority class samples in each subset with the original minority class samples, and we can find the following relationship between $Q$ and the imbalance ratio (IR).

$$Q = \left\lfloor \frac{n_1}{n_2} \right\rfloor \leq IR \tag{16}$$

where $\lfloor \ \rfloor$ denotes rounding down, so that the number of subsets divided among different imbalanced data sets is correlated with IR.

Secondly, after obtaining $Q$ balanced training sets though the above subsets division and fusion (D&F) method, the balanced training sets are input to the *L*-layer DSEN-LG to train the network and obtain the deep envelope training sets, then the deep envelope test set can be obtained via trained DSEN-LG. Finally, the deep envelope training and test sets are used for classification model training and prediction, and a voting mechanism is adopted to determine the final prediction results of the test samples. The specific flowchart can be seen in the figure 3.

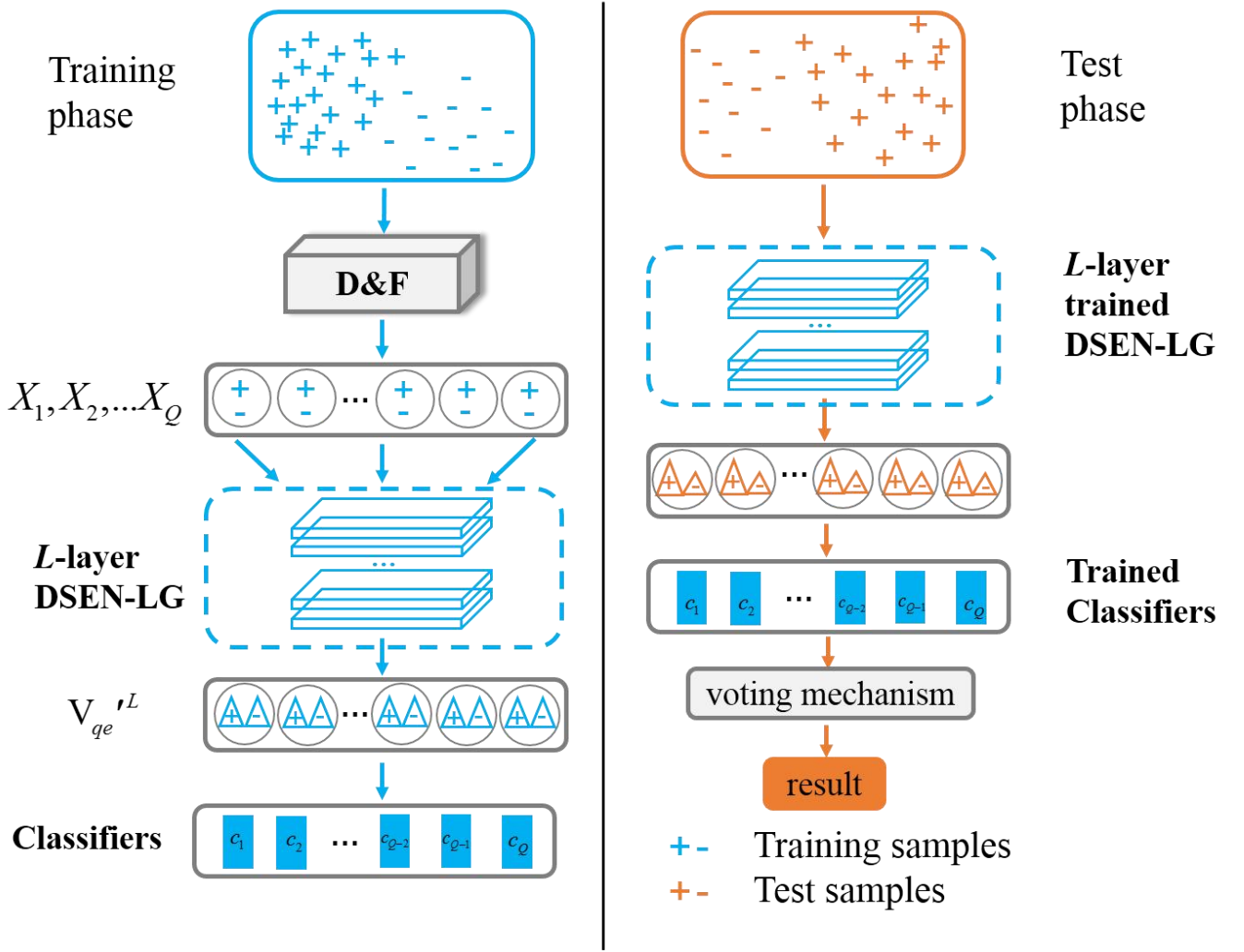

Fig.3.Whole DSEN-LGIE algorithm

## 3. Experimental results and analysis

### 3.1. Experiment conditions

  Forty-four popular publicly available datasets are employed to evaluate the proposed algorithm. The MlFCM and simple ensemble method with bagging mechanism are compared to the proposed method. Additionally, another two groups of ensemble algorithms have also been adopted as the comparison algorithms. The first group comprises seven classical ensemble algorithms:
RUSBoost(RBO)[13],SMOTEBoost(SBO)[14],UnderBagging(UBAG)[15],SMOTEBagging(SBAG)[18],BalancedBagging(BBAG)[26],EasyEnsemble[27],BalanceCascade[27].Each one of these methods represents a distinct combination of an ensemble method (e.g., bagging, boosting and hybrid method). For comparison with more sophisticated algorithms, the second group uses four state-of-the-art ensemble algorithms: CBIS[32], SPE[33], HOEC[34],HD-Ensemble[35]. All the ensemble algorithms are the most relevant and newest. Besides, the diversity and performance on the base classifier of the proposed algorithm are also analyzed.

  Since the most relevant ensembles in this field are decision trees [32-35], decision tree C4.5 was adopted as the base classifier in the experiment. 5-fold cross validation procedure (5-CV) was adopted and the 5-CV procedure was repeated 10 times on every experimental dataset to eliminate the effect of randomness.

### 3.1.1. Datasets

Some imbalanced datasets from the KEEL[47], UCI[48] and Libsvm(https://www.csie.ntu.edu.tw/cjlin/libsvmtools/datasets/) databases are used, and Table 3 gives the information of these datasets, including the number of features, samples, minority class samples and majority class samples, and the imbalance ratio (IR). These datasets are widely used [32-35]. Among these datasets, some have multiple categories and several of these categories are treated as majority class and several of them as minority class. For example, Glass016vs2 is from Glass, where the classes 0,1,6 belong to the majority class and the class 2 belongs to minority class. Winequality-red-8vs6-7 is from Red Wine Quality, where classes 6,7 belong to the majority class and the class 8 belongs to minority class.

Table3. Characteristics of 44 imbalanced datasets

|    | Dataset                 | features | samples | minority | majority | IR    |
|----|-------------------------|----------|---------|----------|----------|-------|
| 1  | Iris0                   | 4        | 150     | 50       | 100      | 2     |
| 2  | Glass0                  | 9        | 214     | 70       | 144      | 2.06  |
| 3  | Vertebral               | 6        | 310     | 100      | 210      | 2.1   |
| 4  | Haberman                | 3        | 306     | 81       | 225      | 2.78  |
| 5  | Vehicle1                | 18       | 846     | 217      | 629      | 2.9   |
| 6  | Blood-transfusion       | 4        | 748     | 178      | 570      | 3.2   |
| 7  | Ecoli1                  | 7        | 336     | 77       | 259      | 3.36  |
| 8  | New-thyroid1            | 5        | 215     | 35       | 180      | 5.14  |
| 9  | Ecoli2                  | 7        | 336     | 52       | 284      | 5.46  |
| 10 | Glass6                  | 9        | 214     | 29       | 185      | 6.38  |
| 11 | Yeast3                  | 8        | 1484    | 163      | 1321     | 8.10  |
| 12 | Ecoli3                  | 7        | 336     | 35       | 306      | 8.6   |
| 13 | Page-blocks0            | 10       | 5472    | 559      | 4913     | 8.79  |
| 14 | Yeast-2-vs-4            | 8        | 514     | 51       | 463      | 9.08  |
| 15 | Yeast-0-5-6-7-9-vs-4    | 8        | 528     | 51       | 477      | 9.35  |
| 16 | Vowel0                  | 10       | 988     | 90       | 898      | 9.98  |
| 17 | Glass016vs2             | 9        | 192     | 17       | 175      | 10.29 |
| 18 | Ecoli-0-1-4-7_vs_2-3-5-6| 7        | 336     | 29       | 307      | 10.59 |
| 19 | climate                 | 18       | 540     | 46       | 494      | 10.7  |
| 20 | Glass2                  | 9        | 214     | 17       | 197      | 11.59 |
| 21 | german                  | 24       | 324     | 24       | 300      | 12.5  |
| 22 | Shuttle-c0-vs-c4        | 9        | 1829    | 123      | 1706     | 13.87 |
| 23 | Yeast-1-vs-7            | 8        | 459     | 30       | 429      | 14.3  |
| 24 | Ecoli4                  | 7        | 336     | 20       | 316      | 15.8  |
| 25 | Page-blocks-1-3-vs-4    | 10       | 472     | 28       | 444      | 15.86 |
| 26 | Dermatology-6           | 34       | 358     | 20       | 338      | 16.9  |
| 27 | svmguide3               | 22       | 312     | 16       | 296      | 18.5  |
| 28 | Yeast-1-4-5-8-vs-7      | 8        | 693     | 30       | 663      | 22.1  |
| 29 | Yeast4                  | 8        | 1484    | 51       | 1433     | 28.10 |
| 30 | Winequality-red-4       | 11       | 1599    | 53       | 1546     | 29.17 |
| 31 | Yeast-1-2-8-9-vs-7      | 8        | 947     | 30       | 917      | 30.57 |
| 32 | Abalone-3_vs-11         | 8        | 502     | 15       | 487      | 32.47 |
| 33 | Yeast5                  | 8        | 1484    | 44       | 1440     | 32.73 |
| 34 | Ozone-onehr             | 72       | 2536    | 73       | 2463     | 33.74 |
| 35 | kr-vs-k-three_vs_eleven | 6        | 2935    | 81       | 3854     | 35.23 |
| 36 | Abalone-21_vs_8         | 8        | 581     | 14       | 567      | 40.5  |
| 37 | Yeast6                  | 8        | 1484    | 35       | 1449     | 41.4  |

| 38 | Winequality-white-3_vs_7 | 11 | 900 | 20 | 880 | 44 |
| 39 | Winequality-red-8vs6-7 | 11 | 855 | 18 | 837 | 46.5 |
| 40 | kr-vs-k-zero_vs_eight | 6 | 1460 | 27 | 1433 | 53.07 |
| 41 | Shuttle-2_vs_5 | 9 | 3316 | 49 | 3267 | 66.67 |
| 42 | kddcup-buffer_overflow_vs_back | 41 | 2233 | 30 | 2203 | 73.43 |
| 43 | kr-vs-k-zero_vs_fifteen | 6 | 2193 | 27 | 2166 | 80.22 |
| 44 | kddcup-rootkit-imap_vs_back | 41 | 2225 | 22 | 2203 | 100.14 |

### 3.1.2. Evaluation metrics and nonparametric statistical tests

To assess the performance of the methods, this paper used AUC[49], F-measure (F-M) [50], G-mean (G-M) [51], Matthews correlation coefficient (Mcc) [52] criteria, and calculated the average results of AUC, F-M, G-M, Mcc of each method on the dataset. The calculation formula of the four criteria are as follows.

$$\text{AUC} = (\text{Sen} + \text{Spe})/2 \tag{17}$$

$$\text{F-M} = 2 \times \frac{\text{Pre} \times \text{Rec}}{\text{Pre} + \text{Rec}} \tag{18}$$

$$\text{G-M} = \sqrt{\text{Sen} \times \text{Spe}} \tag{19}$$

$$\text{Mcc} = \frac{\text{TP} \times \text{TN} - \text{FP} \times \text{FN}}{\sqrt{(\text{TP}+\text{FP})(\text{TP}+\text{FN})(\text{TN}+\text{FP})(\text{TN}+\text{FN})}} \tag{20}$$

where sensitivity(sen), specificity(spe), precision(pre) and recall(rec) are calculated as follows

$$\text{Sen} = \text{TP}/(\text{TP}+\text{FN}) \tag{21}$$

$$\text{Spe} = \text{TN}/(\text{TN}+\text{FP}) \tag{22}$$

$$\text{Pre} = \text{TP}/(\text{TP}+\text{FP}) \tag{23}$$

$$\text{Rec} = \text{TP}/(\text{TP}+\text{FN}) \tag{24}$$

TP is for true positive, FP is for false positive, TN is for true negative and FN is for false negative. Moreover, the nonparametric statistical test methods were adopted such as Friedman test [53] to detect statistical differences between all the methods. In the present study, a Holm post hoc test procedure [54] can be performed to determine the differences between algorithms and α =0.05 was considered the level of significance.

### 3.1.3. Parameter setting

For the proposed method, three parameters need to be determined before running the learning procedure: (1) $K$, which means the number of nearest neighbor samples for SNC, (2) $L$, which is used to determine the number of layers for DSEN-LG, (3) $Q$, as defined in Eq.(16), which determines the number of subsets. We set $K=3, L=3, Q=\lfloor IR \rfloor$, and the Gaussian kernel function was used in the present study. All the results in experiment were obtained under this setting.

### 3.2. Verification of DSEN-LG by ablation method

To demonstrate the effectiveness of deep envelope sample obtained by DSEN-LG, ablation method was adopted to compare the proposed algorithm with the MlFCM and simple bagging method called 'Bgging+None'. MlFCM means the original dataset is clustered by MlFCM. Six datasets represented two types of datasets are chosen (e.g.,high- and low-IR), including Ecoli1,Ecoli2,Ecoli3,Yeast-1-4-5-8-vs-7,Yeast5,Winequality-white-3_vs_7. Table 4 records the results.

Table 4. Ablation method for the proposed method

| Dataset | Ecoli1 | | | | Ecoli2 | | | |
|---|---|---|---|---|---|---|---|---|
| Measure | AUC | F-M | G-M | Mcc | AUC | F-M | G-M | Mcc |
| Proposed method | **0.9246** | **0.8041** | **0.9209** | **0.8047** | **0.9362** | **0.8278** | **0.9275** | **0.8201** |
| Bagging+None | 0.8588 | 0.6934 | 0.8493 | 0.6183 | 0.7131 | 0.3984 | 0.6616 | 0.3194 |
| MlFCM | 0.8185 | 0.6482 | 0.8119 | 0.5487 | 0.7433 | 0.4242 | 0.7086 | 0.3553 |
| Dataset | Ecoli3 | | | | Yeast-1-4-5-8-vs-7 | | | |
| Measure | AUC | F-M | G-M | Mcc | AUC | F-M | G-M | Mcc |
| Proposed method | **0.9569** | **0.7337** | **0.9549** | **0.7430** | **0.7959** | **0.1804** | **0.7205** | **0.2415** |
| Bagging+None | 0.8628 | 0.5022 | 0.8579 | 0.4937 | 0.5838 | 0.1020 | 0.5514 | 0.0702 |
| MlFCM | 0.7815 | 0.3569 | 0.7569 | 0.3468 | 0.5185 | 0.0858 | 0.4388 | 0.0170 |
| Dataset | Yeast5 | | | | Winequality-white-3_vs_7 | | | |
| Measure | AUC | F-M | G-M | Mcc | AUC | F-M | G-M | Mcc |
| Proposed method | **0.9802** | **0.9437** | **0.9751** | **0.9300** | **0.9263** | **0.3539** | **0.9199** | **0.4685** |
| Bagging+None | 0.9490 | 0.3817 | 0.9475 | 0.4602 | 0.7732 | 0.1048 | 0.7589 | 0.1691 |
| MlFCM | 0.8639 | 0.1849 | 0.8529 | 0.2722 | 0.6799 | 0.1088 | 0.6482 | 0.1314 |

From Table 4, the proposed algorithm shows a large improvement in performance on all four metrics compared to MlFCM and bagging+None method. This indicates samples generated through DSEN-LG network are of high quality and very effective. In terms of the four criteria, the proposed method is better than the Bagging+None. It means that the multi-layer clustering can obtain new samples with hierarchical structure information, and the samples have good complementarity, which are helpful imbalance learning. In terms of the four criteria, the proposed method is better than the MIFCM. It means that the LGSCM can well explore the structural information of samples, thereby enhancing the consistency of the interlayer samples of MIFCM.

Besides, inspired by the Kappa-AUC diagram [55] and Kappa-error diagram [56], the Kappa-AUC, F-M, G-M and Mcc diagrams are designed. These diagrams aim to analyze the diversity and performance of base classifiers in the ensemble system. Small kappa values and high AUC, F-M, G-M, Mcc values indicate the base classifiers in the ensemble system have high diversity and excellent classification performance. Figure 4 displays the diversity and corresponding metrics AUC, F-M,G-M, Mcc on Ecoli3 and Yeast-1-4-5-8-vs-7 obtained using proposed method, BalancedBagging, SMOTEBagging and UnderBagging.

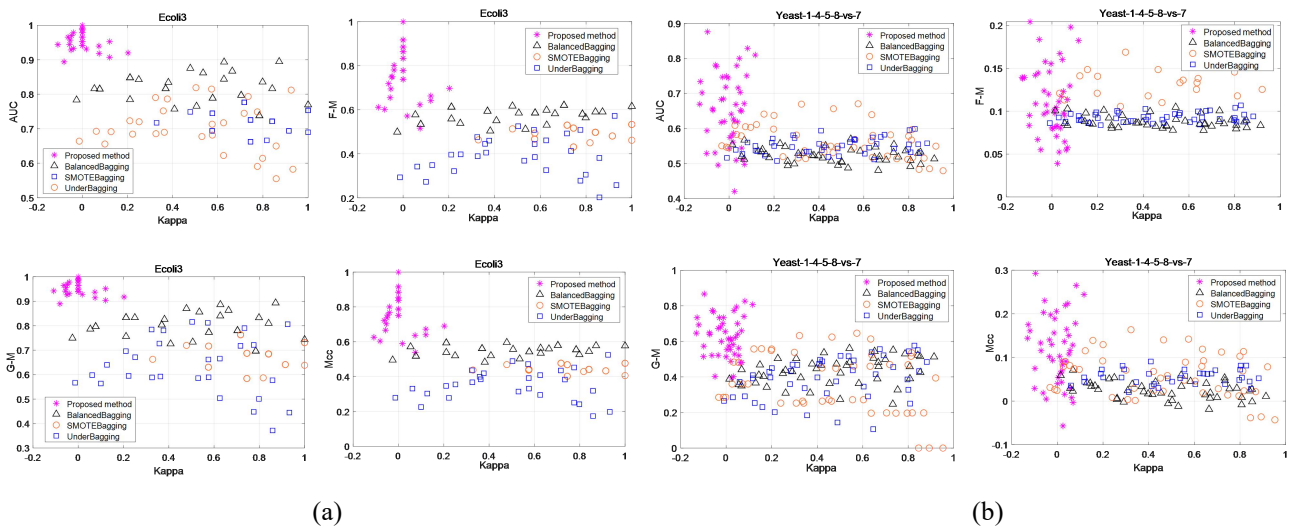

Fig.4. Diversity and performance analysis of base classifiers:(a) diversity and performance analysis for Ecoli3, (b) diversity and performance analysis for Yeast-1-4-5-8-vs-7.

In the Fig.4, it can be seen the points obtained by the proposed algorithm are located in the upper left corner of the figure. It means the kappa values are smaller and AUC, F-M, G-M and Mcc values are higher with the proposed algorithm, indicating the base classifiers of the proposed algorithm have higher diversity and higher performance than other imbalanced ensemble methods. The results means that the proposed DSEN-LG network can obtain new samples with better diversity, which is helpful for subsequent classification, compared to other relevant methods. In general, the proposed algorithm is competitive.

## 3.3. Algorithm comparison

### 3.3.1. Comparison with classical imbalanced ensemble methods

In this experiment, the proposed DSEN-LGIE was evaluated against 7 competitive methods. The brief description of the algorithms can be found in the beginning of the section 3.1. Tables 5 lists the average AUC, F-M, G-M and Mcc values obtained by performing $5 \times 10$ CV procedure on each dataset with different ensemble methods. This section shows the results for two datasets, and the Appendix section gives the complete results of Table 5. The best results are shown in bold style.

Table 5. Comparison results of the ensemble methods on 44 experimental datasets

| Dataset | Measure | RBO | SBO | UBAG | SBAG | BBAG | EasyEnsemble | BalanceCascade | Proposed method |
|---|---|---|---|---|---|---|---|---|---|
| Vertebral | AUC | 0.7400 | 0.7457 | 0.8240 | 0.8036 | 0.8264 | 0.7962 | 0.7843 | **0.8398** |
|  | F-M | 0.6432 | 0.6533 | 0.7535 | 0.7318 | 0.7578 | 0.7121 | 0.7015 | **0.7841** |
|  | G-M | 0.7304 | 0.7319 | 0.8219 | 0.8004 | 0.8245 | 0.7939 | 0.7806 | **0.8298** |
|  | Mcc | 0.4769 | 0.5096 | 0.6311 | 0.6026 | 0.6376 | 0.5599 | 0.5559 | **0.7145** |
| Haberman | AUC | 0.5329 | 0.5741 | 0.5947 | 0.5200 | 0.5889 | 0.5606 | 0.5195 | **0.6694** |
|  | F-M | 0.3050 | 0.4062 | 0.4301 | 0.3040 | 0.4216 | 0.4010 | 0.3404 | **0.5091** |
|  | G-M | 0.4755 | 0.5681 | 0.5882 | 0.4675 | 0.5788 | 0.5552 | 0.5065 | **0.6626** |
|  | Mcc | 0.0624 | 0.1367 | 0.1731 | 0.0497 | 0.1699 | 0.1104 | 0.0362 | **0.3141** |

The experimental results in Table 5 present an overwhelming improvement of DSEN-LGIE over the other imbalanced ensemble methods on all four criteria. In particular, when considering AUC and G-M as the performance criteria, it is observable the method proposed in this paper provided the best performance on 37 and 36 datasets respectively, and never showed the worst performance on any dataset. For F-M and Mcc, the proposed method provided the best performance on 29 and 32 datasets respectively. Thus, DSEN-LGIE perform best in most imbalanced datasets. For example, DSEN-LGIE obtains the highest AUC value, F-M, G-M and Mcc on the high-IR dataset such as Winequality-white-3_vs_7. The result means that the proposed algorithm is better than existing classical imbalance ensemble methods apparently. The possible reason is that the samples obtained by DSEN-LG have higher quality and more classification ability.

Assuming the first rank for the method with the best performance and the eighth rank for the method with the worst performance, so for AUC, F-M, G-M and Mcc, we can calculate and analyze the average ranks of each method on the experimental datasets. Table 6 gives the average ranks of AUC, F-M, G-M and Mcc of each method on the 44 datasets and Figure 5 visualizes the results.

Table 6. Average ranks of all compared ensemble methods

|  | AUC | F-M | G-M | Mcc |
|---|---|---|---|---|
| DSEN-LGIE | **1.591** | **2.500** | **1.727** | **2.273** |
| RBO | 6.272 | 5.500 | 6.273 | 5.750 |
| SBO | 5.864 | 4.182 | 5.773 | 4.386 |
| UBAG | 3.318 | 4.659 | 3.273 | 4.591 |
| SBAG | 6.250 | 3.523 | 6.341 | 3.545 |
| BBAG | 4.023 | 4.864 | 4.045 | 5.023 |
| EasyEnsemble | 3.591 | 5.045 | 3.545 | 5.045 |
| BalanceCascade | 3.864 | 4.477 | 3.795 | 4.318 |

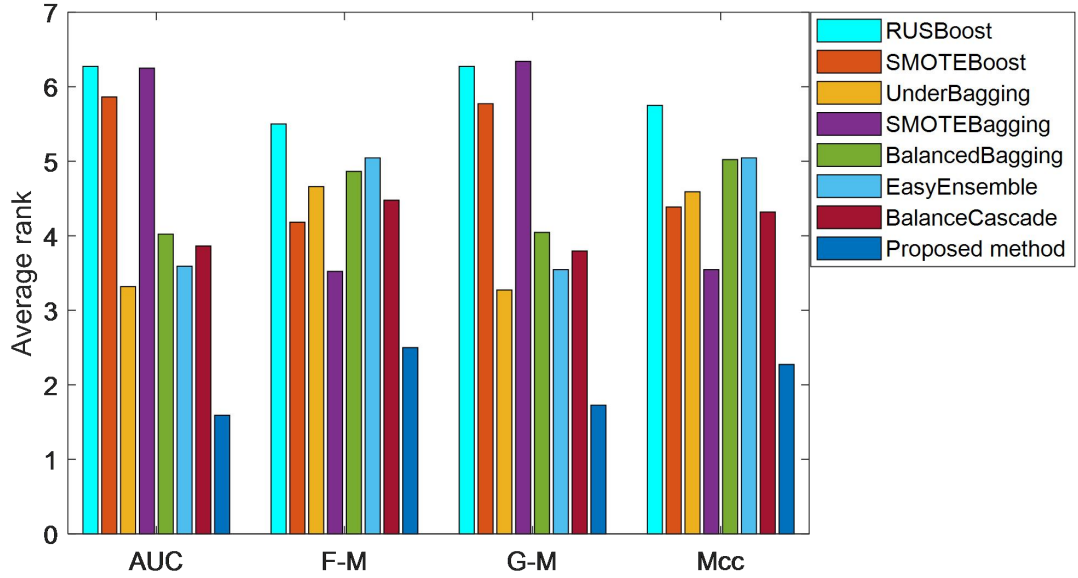

Fig.5. Average ranks of all compared ensemble methods

Table 6 shows the proposed DSEN-LGIE method achieves the lowest average ranks 1.591, 2.500, 1.727 and 2.273 on the AUC, F-M, G-M and Mcc , respectively, so the performance of proposed method is the best. It can be also seen the average ranks of the other methods on the four metrics are clearly different from the DSEN-LGIE in Figure 5, suggesting the excellence performance of DSEN-LGIE has maintained considerable consistency throughout various metrics.

To further evaluate the validity of the DSEN-LGIE, the Holm's test was employed. In the test, we took the proposed method as the control method, and analyzed whether there were the statistically significant differences with other imbalanced ensemble methods in terms of average ranks. the results of Holm's test are recorded in Table 7.

Table 7. Results of Holm's test

| $\alpha_{0.05}$ | AUC | | F-M | |
|---|---|---|---|---|
| | Method | p-value | Method | p-value |
| 0.0073 | RUSBoost | 4.39E-28 | RUSBoost | 8.30E-10 |
| 0.0085 | SMOTEBagging | 7.22E-28 | EasyEnsemble | 1.54E-07 |
| 0.0102 | SMOTEBoost | 2.92E-24 | BalancedBagging | 1.03E-06 |
| 0.0127 | BalancedBagging | 1.20E-09 | UnderBagging | 7.60E-06 |
| 0.0169 | BalanceCascade | 1.19E-08 | BalanceCascade | 3.98E-05 |
| 0.0253 | EasyEnsemble | 4.58E-07 | SMOTEBoost | 4.54E-04 |
| 0.05 | UnderBagging | 1.21E-05 | SMOTEBagging | 0.0320 |
| $\alpha_{0.05}$ | G-M | | Mcc | |
| | Method | p-value | Method | p-value |
| 0.0073 | SMOTEBagging | 4.76E-27 | RUSBoost | 8.15E-13 |
| 0.0085 | RUSBoost | 2.05E-26 | EasyEnsemble | 7.29E-09 |

| 0.0102 | SMOTEBoost | 6.89E-22 | BalancedBagging | 9.53E-09 |
| 0.0127 | BalancedBagging | 8.40E-09 | UnderBagging | 1.11E-06 |
| 0.0169 | BalanceCascade | 2.42E-07 | SMOTEBoost | 8.44E-06 |
| 0.0253 | EasyEnsemble | 5.13E-06 | BalanceCascade | 1.60E-05 |
| 0.05 | UnderBagging | 9.96E-05 | SMOTEBagging | 0.0068 |

From Table 7, It's obvious that all the hypothesis of equivalence have been rejected, indicating the proposed method DSEN-LGIE performs better than the other 7 imbalanced ensemble methods in significance level. Overall, the results in Tables 5-7 show that DSEN-LGIE has a remarkable improvement compared to other imbalanced ensemble methods.

### 3.3.2. Comparison with the state-of-the-art imbalanced ensemble methods

Four state-of-the-art imbalanced ensemble methods were selected for comparison, namely CBIS [32], SPE[33], HOEC[34], HD-Ensemble[35]. The source codes of these methods are not publicly available, except for the SPE method, and their learning process involves multiple parameters to be set. To avoid biased conclusions due to the implementation of the algorithms and the setting of parameters, we extracted the experimental results of CBIS, HOEC and HD-ensemble reported in their original papers. We ran the SPE's code and compare its results with the proposed method. The parameters setting in DSEN-LGIE is based on default for fair comparison. Table 8 records comparison results in this section, and the Appendix section gives the complete results of Table 8.

Table 8. The comparison results between CBIS, HD-Ensemble, HOEC, SPE and DSEN-LGIE

| Dataset | Ecoli2 | | | | Glass6 | | | |
|---|---|---|---|---|---|---|---|---|
| Measure | AUC | F-M | G-M | Mcc | AUC | F-M | G-M | Mcc |
| CBIS | 0.9340 | -- | -- | -- | 0.9340 | -- | -- | -- |
| HD-Ensemble | -- | -- | -- | -- | -- | -- | -- | -- |
| HOEC | 0.9128 | -- | -- | -- | -- | -- | -- | -- |
| SPE | 0.8992 | 0.8067 | 0.8938 | 0.7787 | 0.9164 | 0.8300 | 0.9130 | 0.8074 |
| DSEN-LGIE | 0.9362 | 0.8279 | 0.9276 | 0.8201 | 0.9813 | 0.9591 | 0.9796 | 0.9577 |

The comparisons in Table 8 clearly demonstrated that the proposed DSEN-LGIE provide better performance in terms of the four metrics, suggesting that DSEN-LGIE outperforms the four methods.

## 4. Discussion and Conclusions

Existing imbalanced learning methods always are based on original samples and ignore the structure information among samples. Besides, the structure information among samples includes local and global structure information. Therefore, it is necessary and challenging to explore local and global structure information among samples for improving efficiency of imbalanced learning.

To solve the problem, an imbalanced ensemble method based on DSEN-LG (called DSEN-LGIE)is proposed here. First, the deep sample envelope pre-network (DSEN) is designed to mine structure information among samples, including sample neighborhood concatenation (SNC) and deep envelope sample generation based on multilayer FCM (MlFCM). Then, the LGSCM is proposed to enhance the consistency of the interlayer samples distribution. Next, the DSEN and LGSCM are put together to form the final deep sample envelope network – DSEN-LG. After that, base classifiers are applied on the layers of deep samples respectively. Finally, the predictive results from base classifiers are fused through bagging ensemble mechanism to get the final result.

Over 10 ensemble methods are adopted as the comparison algorithms to evaluate the performance of the proposed algorithm. The comparison results on over 40 imbalanced datasets show that the performance of proposed ensemble algorithm is better than other classical and state-of-the-art imbalanced learning approaches. As seen from Tables 5-7, the proposed algorithm achieves the best performance over classical algorithms significantly. From Tables 8-9, it is also observed the proposed method outperforms the relevant algorithms significantly. Specifically, the proposed method provided the best performance on 37 ,36 ,29 and 32 datasets for AUC, G-M, F-M and Mcc in Table 5 and has the lowest mean ranking on the four metrics at 1.591, 2.500, 1.727 and 2.273 in Table 6. In Table 7, It's obvious that all the hypothesis of equivalence has been rejected. Similarly, compared with the CBIS, HD-Ensemble, HOEC and SPE, proposed DSEN-LGIE provide better performance in term of AUC, F-M, G-M and Mcc in Table 8 and is significantly superior to those four algorithms in Table 9.

Based on these results, several conclusions can be drawn. Firstly, the deep envelope sample learning can explore the structure information of original samples effectively and is helpful to obtain representative samples. Secondly, the deep envelope samples can be better than existing sampling method apparently. Third, deep clustering is designed by combining multilayer clustering and deep transformation, and is effective to realize the deep envelope learning. Finally, domain adaptation is helpful for improving the deep clustering by exploring the local and global structure information of samples.

Besides, seen from the results, the proposed method is more appropriate for processing datasets with high IR than the four state-of-the-art methods. For example, the kinds of datasets include the Glass016vs2, Yeast-1-4-5-8-vs-7, Yeast-1-vs-7 and Yeast5. Conversely, for datasets with low IR like Glass 0 and vehicle1, we can observe in Table 8 that the performance of proposed algorithm is slightly worse or not better apparently. The possible reason is that the number of base classifiers Q of high-IR dataset is higher than that of low-IR dataset, and fewer base classifiers are not conducive to model classification.

Although the proposed algorithm achieved promising results, future works still remain. For example, more clustering algorithms, more datasets and more domain adaptation methods can be considered for further verification in the near future.

## Acknowledgments

The authors would like to thank the editor and reviewers for their valuable comments and suggestions. The authors would also like to thank those individuals or institutions that have provided data support for this research. This work was supported in part by the National Natural Science Foundation of China (NSFC) under grant 61771080 and U21A20448, the Key Project of Technology Innovation and Application Development in Chongqing (cstc2019jsxz-mbdxX0050),the Natural Science Foundation of Chongqing (cstc2020jscx-msxm0369, cstc2020jcyj-msxmX0100, cstc2020jscx-fyzx0212, cstc2020jcyj-msxmX0523, cstc2020jscx-gksbx0009, and The Chongqing Social Science Planning Project (2018YBYY133), and the fund of Sichuan (21ZDYF3646).

## Declarations of interest

None.

## Consent for publication

Not applicable.

## Data availability

The codes and data can be found in (Github): https://github.com/leaphan/ensemble .